\documentclass[a4paper,fleqn]{cas-sc}

\usepackage[authoryear,longnamesfirst]{natbib}

\usepackage{graphicx}
\usepackage{lscape}
\usepackage[aboveskip=0pt,belowskip=-4pt]{caption}
\usepackage{subfig}
\usepackage{float}
\usepackage{adjustbox}
\usepackage{booktabs}
\usepackage{enumitem}

\usepackage{algorithm}
\usepackage{algpseudocode}

\usepackage{amsmath}
\usepackage{bbm}          
\usepackage{gensymb}

\usepackage{xcolor}
\usepackage{colortbl}

\newcommand{\mypar}[1]{\noindent\vspace{0.15em}\textbf{#1}~}

\def\tsc#1{\csdef{#1}{\textsc{\lowercase{#1}}\xspace}}
\tsc{WGM}
\tsc{QE}
\tsc{EP}
\tsc{PMS}
\tsc{BEC}
\tsc{DE}

\begin{document}

\let\WriteBookmarks\relax
\def\floatpagepagefraction{1}
\def\textpagefraction{.001}

\shorttitle{Federated Vision Transformer with Adaptive Focal Loss for Medical Image Classification}

\shortauthors{Zhao et al.}

\title [mode=title]{Federated Vision Transformer with Adaptive Focal Loss for Medical Image Classification}

\fntext[fn1]{Xinyuan Zhao, Yihang Wu and Ahmad Chaddad are contributed equally to this work.}
\author[add1]{Xinyuan~Zhao}
\author[add1]{Yihang~Wu}
\author[add1,add2]{Ahmad Chaddad \corref{cor1}}[orcid=0000-0003-3402-9576]
\ead{ahmadchaddad@guet.edu.cn}
\author[add3, add4]{Tareef~Daqqaq}
\author[add5,add6]{Reem~Kateb}



\address[add1]{School of Artificial Intelligence, Guilin University of Electronic Technology, Guilin, China, 541004}
\address[add2]{The Laboratory for Imagery, Vision and Artificial Intelligence, École de Technologie Supérieure, Montreal, Canada, H3C 1K3}
\address[add3]{College of Medicine, Taibah University, Al Madinah, Saudi Arabia, 42353}
\address[add4]{Department of Radiology, Prince Mohammed Bin Abdulaziz Hospital, Ministry of National Guard Health Affairs, Al Madinah, Saudi Arabia,  42324}
\address[add5]{College of Computer Science and Engineering, Taibah University, Madinah, Saudi Arabia, 42353}
\address[add6]{College of Computer Science and Engineering, Jeddah University, Jeddah, Saudi Arabia, 23445}


\begin{abstract}
While deep learning models like Vision Transformer (ViT) have achieved significant advances, they typically require large datasets. With data privacy regulations, access to many original datasets is restricted, especially medical images. Federated learning (FL) addresses this challenge by enabling global model aggregation without data exchange. However, the heterogeneity of the data and the class imbalance that exist in local clients pose challenges for the generalization of the model. This study proposes a FL framework leveraging a dynamic adaptive focal loss (DAFL) and a client-aware aggregation strategy for local training. Specifically, we design a dynamic class imbalance coefficient that adjusts based on each client's sample distribution and class data distribution, ensuring minority classes receive sufficient attention and preventing sparse data from being ignored. To address client heterogeneity, a weighted aggregation strategy is adopted, which adapts to data size and characteristics to better capture inter-client variations. The classification results on three public datasets (ISIC, Ocular Disease and RSNA-ICH) show that the proposed framework outperforms DenseNet121, ResNet50, ViT-S/16, ViT-L/32, FedCLIP, Swin Transformer, CoAtNet, and MixNet in most cases, with accuracy improvements ranging from 0.98\% to 41.69\%. Ablation studies on the imbalanced ISIC dataset validate the effectiveness of the proposed loss function and aggregation strategy compared to traditional loss functions and other FL approaches. The codes can be found at: \url{https://github.com/AIPMLab/ViT-FLDAF}.

\end{abstract}


\maketitle 

\section{Introduction}
Vision Transformers (ViTs) is a novel promising architectural alternative into a practical backbone for many vision tasks \cite{haruna2025exploring,zhou2025elaformer}. By modeling long‑range interactions through global self‑attention, ViTs can capture contextual dependencies that are critical for medical image interpretation, such as spread of tissue abnormalities or subtle multi‑region cues indicative of disease \cite{yu2023pyramid}. Furthermore, ViT variants and pretraining strategies (e.g., masked image modeling) have increased capacity and sample efficiency, but they remain sensitive to distributional shifts and label scarcity common in clinical datasets \cite{sui2025dm3diff}. In addition, deploying ViTs across healthcare institutions is further prohibited by patient privacy concerns and data silos. Federated learning (FL) offers a practical solution by enabling collaborative model training without centralizing raw images \cite{sultan2025federated,zhao2025fedlgmatch}. However, combining ViTs with FL naively exposes three tightly coupled challenges that are particularly severe in medical imaging: (1) cross‑client heterogeneity (device, protocol and annotation differences) which induces feature and label prior shifts and destabilizes federated averaging \cite{karimireddy2020scaffold,li2020federated}; (2) extreme class imbalance and long‑tailed label distributions \cite{chen2025prototype}; and (3) the interaction between client heterogeneity and class rarity whereby large but internally skewed sites can dominate aggregation and amplify global bias \cite{mohri2019agnostic,li2019fair}. These factors can lead to lower recall score of minority samples, decrease clinical utility, and diminish interpretability that rely on attention maps to localize lesions.

To solve these challenges individually, federated optimization methods (e.g., federated proximal learning, FedProx; stochastic controlled aggregation, SCAFFOLD; federated dynamic adaptation, FedDyn) add proximal or control ‐ variables to stabilize training under heterogeneity \cite{li2020federated,karimireddy2020scaffold,acar2021federated}; long ‐-tailed learning approaches (e.g. focal loss, class‑balanced reweighting, logit adjustment, decoupled training) improve minority performance in centralized settings \cite{Lin2017Focal,cui2019class,cao2019learning,menon2020long,kang2019decoupling}; and emerging ViT–FL studies that use personalization techniques, parameter‑efficient tuning or visual language models to reduce communication or improve adaptation \cite{sun2023fedperfix,lu2023fedclip,zuo2024fedvit}. Yet, two important gaps remain. First, most approaches either treat client imbalance and class imbalance independently or rely on static reweighting that does not adapt to evolving federated dynamics. Second, the potential of attention mechanism used in ViTs as an interpretability signal has been less explored to validate whether distribution‑aware training improves clinical reliability rather than performance metrics. {Furthermore, in standard centralized learning, class distribution is fixed, allowing for pre-calculated static weights. However, in FL, training the global model introduces a challenge: as local models adapt and aggregation weights shift, the global distribution perceived by the model changes dynamically across communication rounds. Static weights fail to capture this temporal information, leading to either under-correction of minority classes or over-fitting to outliers. This motivated us to design a dynamic mechanism that couples prediction (via Focal Loss) with real-time distributional statistics (via the proposed imbalance coefficients) to stabilize federated optimization in non-IID medical image classification.}

In light of these challenges, this study proposes a FL architecture using ViT as the backbone, specifically tailored for the FL framework. We choose ViT as the backbone because it has outperformed CNN models in most tasks \cite{khan2023survey}, and the characteristics of the global attention mechanism better capture the contextual information of the lesion area \cite{parvaiz2023vision}. The key innovation of our approach is a dynamically adjusted focal loss function, which adjusts the loss weights based on the class distribution in each client to address data heterogeneity and class imbalance. By introducing a dynamic imbalance coefficient, our approach ensures that minority class samples receive fair attention, allowing the global model to generalize effectively across diverse client data. This design maintains privacy protection and enhances both model performance. Through validation on three publicly available datasets (ISIC, Ocular Disease, and RSNA-ICH), the experimental results demonstrate performance comparable to state-of-the-art methods. The contributions of this paper are listed as follows.

To summarize, our contributions are:
\begin{itemize}

  \item We formulate a dual‑level imbalance modeling strategy that disentangles client skew and class rarity and injects both into federated optimization and aggregation.
  \item We develop DAFL, a dynamic adaptive focal reformulation that couples prediction hardness with round‑wise distributional statistics to emphasize evolving rarity and difficulty.
  \item We propose a distribution‑aware aggregation rule that reduces the influence of large yet internally skewed clients, improving robustness under severe non‑IID conditions.
  \item We provide extensive empirical evidence across three medical datasets showing improved minority performance, faster federated convergence, and attention‑based interpretability; code and preprocessed splits are released for reproducibility.
\end{itemize}

The rest of the paper is organized as follows. Section \ref{S:2} describes related literature on long‑tailed learning, federated optimization, and ViTs in medical imaging. Section \ref{S:3} details the DAFL formulation and federated protocol. Section \ref{S:4} presents datasets, experimental setup and results, including interpretability analyses. We conclude with limitations and future directions in Section \ref{discussion} and final remarks in Section \ref{conclusion}.

\section{Related Work}\label{S:2}
\noindent\textbf{\textit{Vision transformer.}} The ViT has gained significant attention and been applied across a range of tasks, including image classification, segmentation, and action recognition. For example, CTransCNN \cite{wu2023ctranscnn}, a hybrid model, which combines transformers with CNNs, was proposed. It incorporated a multi-label multi-head attention enhancement module, an information interaction module, and a multi-branch residual module, achieving efficient capture of label and image features. This model demonstrated superior performance compared to existing methods in multi-label medical image classification tasks. Furthermore, in \cite{li2023lvit}, they introduced a novel ViT model, which enhances medical image segmentation by integrating medical text annotations with image data. The model also introduced an Exponential Pseudo Label Iteration mechanism and a Language-Vision Loss to improve the quality of pseudo labels in semi-supervised learning. Furthermore, many ViT variants have been developed to handle imbalanced datasets. 
In \cite{seo2024relaxed}, a dual-fusion method based on the Conformer architecture \cite{gulati2020conformer} was proposed for ocular disease detection. This method combines local features from CNNs and global features from Transformers, with optimizations in both the depth and width of the model. In \cite{xu2023learning}, the authors proposed combining Mask Generation Pretraining and Balanced Binary Cross Entropy (Bal-BCE) to train ViT models, addressing the challenge of long-tailed datasets. Furthermore, in \cite{selvano2023evaluating}, they present a Distillation with No Labels-based self-supervised pretraining ViT model, which is used to classify lung diseases on imbalanced datasets.

\noindent\textbf{\textit{Federated learning.}}  FL effectively avoids the risks of centralized storage and transmission of sensitive data by training models on local devices and only sharing model parameters \cite{gadekallu2023guest}. It is generally classified into three types: horizontal FL, vertical FL, and federated transfer learning \cite{chaddad2023federated}. Taking FedAVG as an example, it updates the global model by aggregating model parameters from each client without the need to share local data \cite{mcmahan2017communication}. To address the specific challenge of \textit{class imbalance} in FL, several optimization strategies have been proposed. In \cite{dai2023tackling}, they addressed class imbalance by leveraging the uniform distribution of class prototypes and class semantic information. In \cite{sarkar2020fed}, focal loss was introduced to the FL framework, outperforming FedProx with reasonable costs. Furthermore, Fed-IT introduced an information-theoretic loss function to address the scattering and gravitational issues of minority classes \cite{hamidi2024fed}, while Fed-CBS proposed a heterogeneous-aware client sampling mechanism to reduce class imbalance \cite{zhang2023fed}. Recent advancements have focused on tackling \textit{statistical heterogeneity} through advanced architectural or data-driven approaches. {To address label-induced heterogeneity, in \cite{collins2021exploiting}, they decouple the model into a shared global representation and personalized local heads, demonstrating robust performance against label skew (FedRep). In \cite{xiong2023feddm}, they incorporate data distillation to generate representative synthetic samples locally to minimize knowledge loss during aggregation (FedDM). To tackle statistical heterogeneity via knowledge transfer, federated multiple classifier aggregation (FedMCA) \cite{zheng2025personalized} employs multiple classifier aggregation and knowledge distillation to mitigate performance degradation. To handle data distribution heterogeneity, federated mixture-of-expert (FtMoE) \cite{liu2025ftmoe} integrates the MoE into a FL framework, employing a gated network to dynamically weight expert opinions based on local data features.}

\noindent\textbf{\textit{ViT in FL for medical imaging.}} The application of ViT in FL has gained significance recently, yet its use remains limited. For example, FedPerfix partial personalized handling of the ViT model improves personalized FL and demonstrates its optimal performance in multiple datasets \cite{sun2023fedperfix}. In \cite{sahoo2022vision}, they proposed a FL framework based on ViT for COVID-19 detection using chest X-ray images. Furthermore, in \cite{zuo2024fedvit}, they explored the use of ViT for federated continuous learning. Their experimental results are superior to the state-of-the-art models using popular federated continuous learning benchmarks. In \cite{nath2025mtmedformer}, they extend ViT to multi-task FL for medical imaging. Specifically, they introduce a ViT to learn task-agnostic features and use task-specific decoders for robust feature extraction. They further design a novel Bayesian federation method for aggregating multi-task imaging models. In \cite{amaizu2025fedvitbloc}, they combine ViT with block chain technique in FL to build a privacy-preserving medical image analysis framework. Using homomorphic encryption and differential privacy mechanisms, their model provides feasible performance while enhances the security and trustworthiness of the FL system.

\noindent\textbf{\textit{Imbalance-aware in federated learning.}}
{While recent methods like FedRep, FedDM, and FtMoE address heterogeneity through architectural decoupling, synthetic data generation, or MoE structures, they often introduce a higher computational overhead or architectural constraints. For instance, FedRep requires maintaining separate local heads, and FedDM involves computationally intensive data synthesis. Unlike these methods, the proposed DAFL offers a unified, optimization-based solution that requires no architectural changes to the ViT backbone. The distinct advantages of DAFL are: 
(i) Dual-level coupling. DAFL mathematically fuses prediction hardness via Focal Loss with real-time distributional statistics via imbalance coefficients into a single objective; (ii) Dynamic Adaptation. Unlike static re-weighting or fixed architectural splits, the coefficients used in DAFL evolve dynamically with the training rounds, capturing the temporal shifts in the global model bias; (iii) Efficiency. DAFL achieves robustness against non-IID data by transmitting a global imbalance parameter with model parameters rather than gradients of multiple experts or synthetic images, preserving both communication efficiency and privacy.}




\section{Methodology}\label{S:3}

\begin{figure}[]
     \centering
\includegraphics[width=1\linewidth]{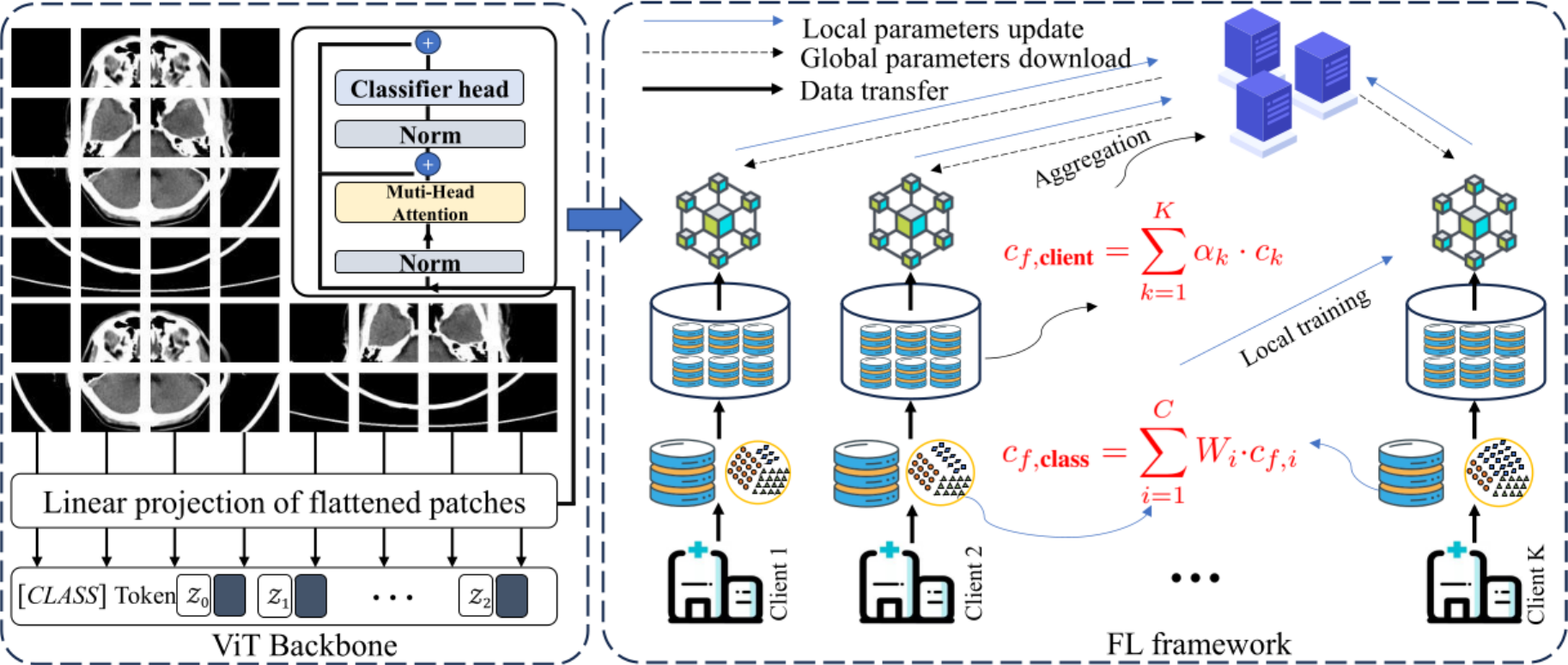}
\caption{Flowchart of our proposed method. The framework integrates a ViT backbone within a federated learning structure. The client-level imbalance parameter (Eq. \ref{eq:client_imbalance}) and the global class-level imbalance parameter (Eq. \ref{eq:class_imbalance}) are calculated based on data distributions to dynamically guide the federated aggregation and local training processes, respectively.}
    \label{figureVIT}
 \end{figure}

In this section, we detail the proposed Federated Learning framework with Dynamic Adaptive Focal Loss (DAFL). Our approach is designed to address the dual challenges of statistical heterogeneity and class imbalance inherent in real-world medical data, particularly within a privacy-preserving federated setting. The methodology begins with a formal problem definition, followed by an overview of the Vision Transformer (ViT) backbone. We then introduce the cornerstone of our method: a dual-level imbalance modeling strategy. Building upon this, we elaborate on the two primary innovations: the Dynamic Adaptive Focal Loss (DAFL) for local client training and a distribution-aware aggregation mechanism for global model updates. The overall architecture and workflow are illustrated in Figure \ref{figureVIT}.

\subsection{Problem formulation}
We address the task of image classification in a FL framework, where a global model is trained across \( K \) clients, each holding a local dataset \( D_k = \{(\mathbf{x}_i, y_i)\}_{i=1}^{N_k} \). Here, $\mathbf{x}_i$ is an input image, and \( y_i \in \{1, 2, \ldots, C\} \) is the class label from \( C \) classes. The global dataset is \( D = \bigcup_{k=1}^K D_k \). The key challenge is data heterogeneity due to non-i.i.d. distributions across clients and class imbalances, where some classes have significantly fewer samples than others. Our goal is to optimize a global model \( f(\mathbf{x}; \Theta) \), parameterized by \( \Theta \), to achieve robust classification performance across all clients and classes. We formulate the optimization problem as minimizing a weighted average of local losses:
 \begin{equation}
    \min_{\Theta} \sum_{k=1}^K \omega_k \mathcal{L}_k(\Theta; D_k),
 \end{equation}
 where \( \mathcal{L}_k \) is the local loss on client \( k \)'s dataset, and \( \omega_k \) is its corresponding aggregation weight. Our framework introduces novel, adaptive formulations for both \(\mathcal{L}_k\) and \(\omega_k\) to specifically counteract the performance degradation caused by data heterogeneity.

\mypar{ViT framework} Unlike CNNs that handle images in a hierarchical fashion, ViT treats an image as a series of flattened patches. It uses a \textit{ transformer encoder} to capture global context, enabling flexible weighting of different image regions depending on the task. This approach allows ViT to dynamically adjust focus on various parts of the image based on the task-specific needs.

For example, ViT divides the image \(x\) (where \(x\) represents the input image) into \(N\) patches for processing. Each patch is of a predefined size, which results in \(N = 16 \times 16\) patches for a standard input image size. Each patch \(p_i\) (where \(p_i\) is the \(i\)-th patch in the set of all patches \(P = \{p_1, p_2, \ldots, p_N\}\)) is flattened into a one-dimensional vector and then linearly projected into a \(D\)-dimensional embedding space using a trainable projection matrix \(E\) (where \(E \in \mathbb{R}^{D \times d_p}\) is the learnable projection matrix and \(d_p\) is the dimensionality of the flattened patch). This projection is represented as:
\begin{equation}
    z_i = E \cdot p_i, \text{ for } i = 1, 2, \ldots, N
\end{equation}

where \(z_i \in \mathbb{R}^{D}\) is the embedded representation of patch \(p_i\).

A special token, [CLASS], is added to the sequence of embedded patches. This token is represented as \(z_0\) and is initialized randomly. The modified sequence is:
\begin{equation}
    Z = \{z_0, z_1, z_2, \ldots, z_N\}
\end{equation}

The [CLASS] token serves as a container to accumulate global information about the image through the self-attention mechanism. The sequence of embeddings \(\mathbf{Z}\) (where \(\mathbf{Z} \in \mathbb{R}^{(N+1) \times D}\)) is processed through multiple layers of a ViT encoder to obtain an encoded feature representation \(\mathbf{F}_{\text{encoded}} \in \mathbb{R}^{(N+1) \times D}\):

\begin{equation}
\mathbf{F}_{\text{encoded}} = \text{ViTEncoder}(\mathbf{Z}; \Theta)
\end{equation}
where \(\Theta\) represents the parameters of the ViT model.

\textit{Input Embedding and Position Encoding:} Each input token \(x_i\) (where \(x_i\) is the \(i\)-th token in the sequence) is first converted into a high-dimensional space vector through an embedding process. This is essential to represent discrete tokens as continuous vectors that the model can process:
\begin{equation}
    \text{Embed}(x_i) = E x_i
\end{equation}

where \(E \in \mathbb{R}^{D \times d_t}\) is the embedding matrix and \(d_t\) is the dimensionality of the token embedding space.

After embedding, positional encoding are added to the embedded vectors. This step injects information about the relative or absolute position of the tokens in the sequence:
\begin{equation}
    z_0 = \text{Embed}(X) + P
\end{equation}

where \(X = (x_1, x_2, \dots, x_n)\) represents the sequence of input tokens, \(P \in \mathbb{R}^{n \times D}\) is the position encoding matrix that varies sinusoidally, and \(n\) is the number of tokens in the sequence.

\textit{Multi-Head Self-Attention:} The multi-head self-attention mechanism allows the model to dynamically focus on parts of the input sequence and understand the relationships between all tokens. It is calculated as follows:
\begin{equation}
    \text{Attention}(Q, K, V) = \text{softmax}\left(\frac{QK^T}{\sqrt{d_k}}\right)V
\end{equation}

where \(Q\), \(K\), and \(V\) are the queries, keys, and values matrices, respectively, computed as linear transformations of the input. Specifically:
\begin{equation}
    Q = z_{\ell-1} W^Q, \quad K = z_{\ell-1} W^K, \quad V = z_{\ell-1} W^V
\end{equation}

Here, \(W^Q, W^K, W^V \in \mathbb{R}^{D \times d_k}\) are the learnable weight matrices, and \(d_k\) is the dimensionality of the key vectors.

The scaled dot-product attention allows each token to interact with every other token in a weighted manner, scaled by the square root of \(d_k\) to avoid overly large dot products. The outputs of individual attention processes are then combined in a multi-head scheme:
\begin{equation}
    \text{MultiHead}(Q, K, V) = \text{Concat}(\text{head}_1, \dots, \text{head}_h)W^O
\end{equation}

where \(W^O \in \mathbb{R}^{h \cdot d_k \times D}\) is the output projection matrix, and \(h\) is the number of attention heads. For each head:
\begin{equation}
    \text{head}_i = \text{Attention}(QW_i^Q, KW_i^K, VW_i^V)
\end{equation}

where \(W_i^Q, W_i^K, W_i^V\) are the weight matrices for the \(i\)-th attention head.

\textit{Add \& Norm:} The output from the multi-head attention is processed further to ensure smooth training and stabilization:
\begin{equation}
    z' = \text{LayerNorm}(z_{\ell-1} + \text{MultiHead}(Q, K, V))
\end{equation}

Layer normalization and residual connections help in training deeper models by alleviating the vanishing gradient problem and encouraging gradient flow through the network.

\textit{Position-wise Feed-Forward Network:} Each position is independently processed by a position-wise feed-forward network, which consists of two linear transformations with a Rectified Linear Unit (ReLU) activation in the middle:
\begin{equation}
    \text{FFN}(x) = \max(0, xW_1 + b_1)W_2 + b_2
\end{equation}

where \(W_1 \in \mathbb{R}^{D \times d_{ff}}\), \(W_2 \in \mathbb{R}^{d_{ff} \times D}\), and \(d_{ff}\) is the dimensionality of the feed-forward network hidden layer. This module enhances the non-linearity of the model's capabilities in processing data. The output is then combined with the input to the module through another residual connection and normalized:
\begin{equation}
    z_\ell = \text{LayerNorm}(z' + \text{FFN}(z'))
\end{equation}

\textit{Output:} The final output \(z_L\) from the last layer of the encoder is typically used as the input to the Transformer decoder in tasks involving language translation or directly for classification tasks. This output encapsulates the complex interactions between all input tokens and effectively represents the entire input sequence.

\noindent\subsection{Dual-Level imbalance modeling}

{The core contribution of the proposed approach is a novel strategy for quantifying data imbalance at two distinct yet complementary levels: the individual client level and the global federation level. This dual-level characterization is important, as a class that is prevalent across the federation may still be underrepresented for a particular client, and conversely, a class that is frequent for a client may be rare globally. Consequently, addressing imbalance at only one of these levels is inadequate for robust and equitable model training in federated settings.}

\noindent{\mypar{Client-Level imbalance} For each client \(k\), we define a client-specific imbalance parameter, \(c_k\) to measure the skewness of its local data distribution. It is calculated as the average of the imbalance ratios across all classes within that client:
\begin{equation}
c_k = \frac{1}{C} \sum_{i=1}^{C} \frac{N_k - n_{k,i}}{n_{k,i} + \epsilon}
\label{eq:client_imbalance}
\end{equation}
where \(N_k\) is the total number of samples at client \(k\), \(n_{k,i}\) is the number of samples for class \(i\) at client \(k\), and \(\epsilon\) is a small constant to prevent division by zero. A higher \(c_k\) value represents a more imbalanced and less representative local dataset. This parameter is a key indicator of data quality or distributional bias for clients and is primarily used to adjust their influence during the global model aggregation phase.}

\noindent{\mypar{Global class-level imbalance} To address the problem of globally rare classes (e.g., a rare disease type across all hospitals), we define a global imbalance parameter for each class \(i\), denoted as \(c_{f,i}\). This parameter is computed by the central server using aggregated statistics from all participating clients:
\begin{equation}
c_{f,i} = \frac{\sum_{k=1}^{K} N_k - \sum_{k=1}^{K} n_{k,i}}{\sum_{k=1}^{K} n_{k,i} + \epsilon}
\label{eq:class_imbalance}
\end{equation}}

{This parameter quantifies the rarity of a class within the entire federated network. It acts as a global signal to guide the local training process. Its primary role is to ensure that globally underrepresented classes receive increased attention, regardless of their local prevalence on any single client.}

\subsection{DAFL: dynamic adaptive focal loss and aggregation}

{Building on the dual-level imbalance modeling, we introduce our complete framework, which synergistically combines a novel loss function for local training with a new strategy for global aggregation. This integrated approach ensures that data imbalance is tackled at both the local training and global aggregation stages.}

\noindent{\mypar{Dynamic adaptive focal loss (DAFL) for local training}
To mitigate the challenges of class imbalance and client heterogeneity during local training, we propose the DAFL. Standard focal loss addresses easy and hard examples, but in FL setting, the importance of a sample depends on its class rarity and the statistical uniqueness of its local training stage. DAFL extends focal loss by incorporating a dynamic coefficient to solve this limitation. For a given sample with true class \(t\) on client \(k\), the loss is defined as:
\begin{equation}
\mathcal{L}_{\text{DAFL}} = - (1 + c_{f,t}) \cdot (1-p_t)^\gamma \cdot \log(p_t)
\label{eq:dafl}
\end{equation}
where \(p_t\) is the predicted probability and \(\gamma\) is a trainable parameter. We set the initial value of $\gamma$ to 2 for all experiments. The key innovation is the \textit{dynamic imbalance coefficient}, \(c_{f,t}\), which we formulate as a weighted combination of two distinct imbalance signals:
\begin{equation}
c_{f,t} = \lambda \cdot c_k + (1-\lambda) \cdot c_{f,i}\big|_{i=t}
\label{eq:dynamic_coeff} 
\end{equation}}
{where $\lambda$ is a balancing parameter. We set $\lambda$ to 0.5 for all experiments.}


\subsection{Federated aggregation with DAFL}
In FL framework, each client trains its local model using its local data and computes its imbalance parameter \( c_k \) based on its class distribution (see Eq. (\ref{eq:client_imbalance})), which reflects the degree of imbalance among the different classes in that client data. We then combine the class information from all clients to obtain a client-level global imbalance parameter using Eq. \ref{eq:class_imbalance}. Based on each client \( c_k \) (or more directly, based on the statistics of each client, which generate a new weight \(\tilde{w}_k\)).

\begin{equation}
w_k = \frac{1}{c_k + \epsilon},
\end{equation}
\vspace{-0.1cm}

Furthermore, to ensure numerical stability, we normalize it as:
\begin{equation}
\omega_k = \frac{w_k}{\sum_{j=1}^{K} w_j}.
\end{equation}

Finally, the global model parameters are updated via weighted aggregation as
\begin{equation}\label{EQ:agg}
\Theta_g \leftarrow \sum_{k=1}^{K} \omega_k \cdot \Theta_k.
\end{equation}

This strategy ensures that clients with relatively balanced data (i.e., lower \( c_k \)) receive higher aggregation weights, thereby enabling the global model to better capture the characteristics of minority classes and overall data. Meanwhile, during local training, we use the class-level imbalance parameter \( c_{f,\text{class}} \) computed at each client to dynamically adjust the adaptive focal loss function, ensuring that the local model obtains sufficient gradient updates even when certain classes are sparse. The global aggregation stage considers the simple averaging of model parameters and uses the information on the client imbalance \( c_{f,\text{client}} \) to generate dynamic weights, achieving adaptive regulation based on data volumes and client-specific characteristics, and ultimately improving the robustness and generalization of the global model in the federated environment.
The detailed steps of the ViT backbone and FL framework used in DAFL are provided in Algorithm \ref{alg:param_calc} and Algorithm \ref{alg:dfl}.

\begin{algorithm}[!ht]\footnotesize
\caption{Federated Learning with Dynamic Adaptive Focal Loss (DAFL)}
\label{alg:dfl}
\begin{algorithmic}[1]
\State \textbf{Server executes:}
\State Initialize global model \(\Theta^0\).
\For{each communication round $t = 1, 2, \ldots, T$}
    \State Server selects a subset of clients \(S_t\).
    \State Server broadcasts \(\Theta^{t-1}\) to selected clients.
    \State \textbf{Client-side computation (in parallel for each client \(k \in S_t\)):}
    \State \quad \(\Theta_k^t \leftarrow \Theta^{t-1}\)
    \State \quad Client \(k\) computes its local imbalance \(c_k\) using Eq. \ref{eq:client_imbalance} and sends it to the server.
    \State \quad \textbf{for} each local epoch \(e = 1, \dots, E\) \textbf{do}
    \State \quad \quad \textbf{for} each batch \((\mathbf{x}, y)\) in \(D_k\) \textbf{do}
    \State \quad \quad \quad Update \(\Theta_k^t\) by descending the gradient \(\nabla \mathcal{L}_{\text{DAFL}}(\Theta_k^t)\) using Eq. \ref{eq:dafl}.
    \State \quad \quad \textbf{end for}
    \State \quad \textbf{end for}
    \State \quad Client \(k\) sends the updated model \(\Theta_k^t\) to the server.
    \State \textbf{Server-side aggregation:}
    \State Server computes global class imbalance \(c_{f,i}\) for all classes \(i\) using Eq. \ref{eq:class_imbalance}.
    \State Server computes aggregation weights \(\{\omega_k\}_{k \in S_t}\) using the received \(c_k\) values.
    \State Server updates the global model: \(\Theta^t \leftarrow Eq .\ref{EQ:agg}\).
\EndFor
\State \textbf{Return:} Final global model \(\Theta^T\).
\end{algorithmic}
\end{algorithm}

\subsection{Theoretical analysis}
{The effectiveness of DAFL framework can be attributed from two optimization perspectives: gradient rectification and aggregation variance reduction.}

\noindent{\mypar{Gradient rectification} The DAFL loss function (Eq. \ref{eq:dafl}) acts as a \textit{gradient scaler}. In standard training, gradients from majority classes often dominate the optimization process. By introducing the adaptive term \((1 + c_{f,t})\), we amplify the gradient magnitude for samples from minority classes or skewed clients. This solves the vanishing gradient problem for underrepresented data, ensuring that their features are learned effectively (Section \ref{discussion}).}

\noindent{\mypar{Variance reduction} In non-i.i.d. settings, the local gradients \(\nabla \mathcal{L}_k\) can vary across clients, leading to high variance in the aggregated global update and unstable convergence. Our distribution-aware aggregation strategy directly addresses this challenge by assigning lower weights (\(\omega_k\)) to clients with high internal skew (\(c_k\)) to decrease the contribution of outlier updates. This prohibits the aggregation to pull the global model away from a generalized solution. This acts as a regularizer on the aggregation process, minimizing the variance of the global update direction and promoting a more stable convergence.}

\begin{algorithm}[!ht]\footnotesize
\caption{Parameter Calculation Logic for DAFL}
\label{alg:param_calc}
\begin{algorithmic}[1]
\State \textbf{Input:} Client datasets \(\{D_k\}_{k=1}^K\), hyperparameter \(\lambda\)
\State \textbf{Output:} Loss coefficient \(c_{f,t}\) for each sample, Aggregation weights \(\{\omega_k\}_{k=1}^K\)

\Statex \textit{//--- Parameter Calculation on Server and Clients ---}
\State \textbf{On each Client \(k\):}
\State \quad For each class \(i\), count local samples \(n_{k,i}\) from \(D_k\).
\State \quad Calculate client-level imbalance \(c_k\) using Eq. \ref{eq:client_imbalance}.
\State \quad Send local counts \(\{n_{k,i}\}_{i=1}^C\) and imbalance \(c_k\) to the server.

\State \textbf{On the Server:}
\State \quad Receive counts and imbalance values from all clients.
\State \quad \textbf{for} each class \(i \in \{1, \dots, C\}\) \textbf{do}
\State \quad \quad Calculate global class-level imbalance \(c_{f,i}\) using Eq. \ref{eq:class_imbalance}.
\State \quad \textbf{end for}
\State \quad Calculate aggregation weights \(\{\omega_k\}_{k=1}^K\) using the received \(c_k\) values.
\State \quad Send the global class imbalance vector \(\{c_{f,i}\}_{i=1}^C\) back to clients.

\Statex \textit{//--- Usage in Local Training ---}
\State \textbf{On each Client \(k\) during local training:}
\State \quad For a sample with true class \(t\):
\State \quad \quad Compute the dynamic imbalance coefficient: 
\State \quad \quad \(c_{f,t} \leftarrow \lambda \cdot c_k + (1-\lambda) \cdot c_{f,i}\big|_{i=t}\)
\State \quad \quad Use \(c_{f,t}\) to compute the \(\mathcal{L}_{\text{DAFL}}\) as per Eq. \ref{eq:dafl}.
\end{algorithmic}
\end{algorithm}


\section{Experiments}\label{S:4}

\subsection{Datasets}

\noindent\mypar{Ocular Disease} It is an ocular disease recognition dataset. These images are sourced from a wide range of datasets, including IDRiD \cite{idrid}, Ocular recognition \cite{ocularrecognition}, High-Resolution Fundus (HRF) \cite{HRF}.

\noindent\mypar{ISIC2019} ISIC2019 is a dataset composed of various skin diseases and skin cancer, containing a total of 7 classes \cite{codella2018skin,combalia2019bcn20000,tschandl2018ham10000,Skin}. Note that this dataset is class imbalanced as the samples in each class varies.

\begin{figure}
    \centering
    \includegraphics[width=0.325\linewidth]{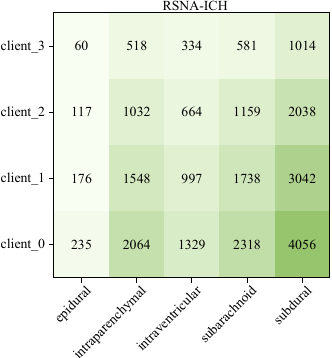} \includegraphics[width=0.32\linewidth]{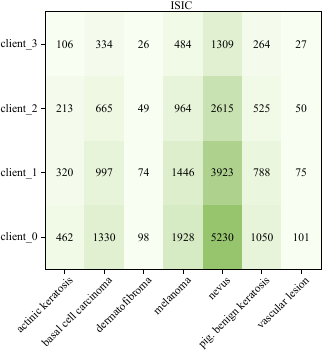} \includegraphics[width=0.325\linewidth]{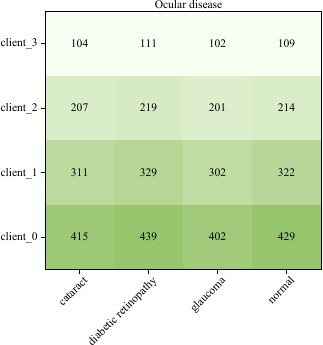} 
    \caption{The number of samples of each client in RSNA-ICH, Ocular disease and ISIC datasets.}
    \label{fig:Client_Samples}
\end{figure}

\noindent\mypar{RSNA-ICH} The RSNA Intracranial Hemorrhage Detection dataset is used to classify acute intracranial \cite{RSNA-ICH} hemorrhages and their five subtypes in CT scans. The dataset consists of de-identified CT scans provided by multiple research institutions and was labeled by over 60 volunteers organized by the American Society of Neuroradiology (ASNR), with more than 25000 examination records. Since most of the images in the dataset come from healthy patients and hemorrhage subtypes are rare, the authors of the data set randomly selected 25000 slices from images containing hemorrhage subtypes and divided them into 80\% for training, 10\% for validation, and 10\% for testing. Slices of the same patient did not appear in splits to avoid patient overlap. Table \ref{table:class_name} reports the details of each class in these datasets, while Figure \ref{fig:Client_Samples} shows the number of samples in each client for these datasets.

\begin{table}[!ht]
\scriptsize
\centering
\caption{Abbreviations with corresponding class names for each dataset. A-E, F-I, and J-P represent RSNA-ICH, Ocular Disease, and ISIC datasets, respectively.}
\setlength{\tabcolsep}{0.42cm}
\renewcommand{\arraystretch}{0.7}
\begin{tabular}{p{0.1cm} p{3.5cm} p{0.1cm} p{2.4cm}}\toprule
 \rowcolor{gray!15}A & Subdural & E & Epidural \\
B & Subarachnoid & F & Normal \\
 \rowcolor{gray!15}C & Intraventricular & G & Glaucoma \\
D & Intraparenchymal & H & Diabetic \\
 \rowcolor{gray!15}I & Cataract & J & Vascular lesion \\
K & Pigmented benign keratosis & L & Nevus \\
 \rowcolor{gray!15}M & Melanoma & N & Dermatofibroma \\
O & Actinic keratosis & P & Basal cell carcinoma \\
\bottomrule
\end{tabular}
\label{table:class_name}
\end{table}

\begin{table}[!ht] \scriptsize
    \centering
    \setlength{\tabcolsep}{0.55cm}
    \caption{Summary of samples used in each client. $C_1$ to $C_{3}$ are the clients while $C_{glo}$ is the global server.}
    \renewcommand{\arraystretch}{0.7}
    \begin{tabular}{ccccc}
    \toprule
    Datasets & $C_1$ & $C_2$ & $C_3$ & $C_{glo}$  \\
    \midrule
   \rowcolor{gray!15} Ocular & 1685 & 1264 & 841 & 587 \\ 
    ISIC & 10163 & 7623 & 5081 & 2552 \\ 
  \rowcolor{gray!15} RSNA-ICH & 10002 & 7501 & 5000 & 2507 \\ 
    \bottomrule
    \end{tabular}
    \label{tab:Samples}
\end{table}

\begin{table*}[ht!]\scriptsize
  \centering
  \caption{{Performance metrics (\%) for classification tasks. \textcolor{cyan}{Cyan} represents the best values. * indicates the federated methods}}
  \renewcommand{\arraystretch}{0.7}  
  \setlength{\tabcolsep}{4.3pt}  
  \begin{tabular}{ccccccc|cccccc|cccccccccc}
    \toprule
    & \multicolumn{3}{c} \textbf{Accuracy} &\multicolumn{3}{c} \textbf{F1} & \multicolumn{3}{c}\textbf{Precision} & \multicolumn{3}{c}\textbf{Recall} &\multicolumn{3}{c} \textbf{Specificity} & \multicolumn{2}{c}\textbf{AUC}\\
    \cmidrule(lr){2-19}  
    \multirow{1}{*}{Model} & \multicolumn{6}{c|}{RSNA-ICH} & \multicolumn{6}{c|}{Ocular Disease} & \multicolumn{6}{c}{ISIC} \\
    \midrule
     \rowcolor{gray!15} DenseNet121 \cite{huang2017densely} & 81.26 & 0.58 & 0.58 & 0.59 & 0.93 & 0.87 & 87.07 & 0.87 & 0.87 & 0.87 & 0.96 & 0.97 & 74.31 & 0.73 & 0.73 & 0.74 & 0.94 & 0.93 \\
  ResNet50 \cite{he2016deep} & 79.53 & 0.76 & 0.80 & 0.73 & 0.93 & 0.94 & 82.21 & 0.82 & 0.82 & 0.82 & 0.94 & 0.96 & 69.09 & 0.67 & 0.67 & 0.69 & 0.93 & 0.90 \\
     \rowcolor{gray!15} ViT-S/16 \cite{rwightman2019timm} & 79.87 & 0.63 & 0.63 & 0.64 & 0.90 & 0.89 & 91.00 & 0.91 & 0.91 & 0.91 & 0.97 & 0.98 & 83.77 & 0.84 & 0.84 & 0.84 & 0.96 & 0.97 \\
   ViT-L/32 \cite{rwightman2019timm} & 82.47 & 0.64 & 0.64 & 0.65 & 0.91 & 0.90 & 92.06 & 0.92 & 0.92 & 0.92 & 0.97 & 0.99 & 85.84 & 0.86 & 0.86 & \textcolor{cyan}{0.86} & 0.97 & 0.97 \\
     \rowcolor{gray!15} $\ast$ FedCLIP$^\dag$ \cite{lu2023fedclip} & 44.67 & 0.44 & 0.47 & 0.44 & 0.90 & 0.63 & 86.42 & 0.86 & 0.86 & 0.86 & 0.86 & 0.90 & 54.91 & 0.59 & 0.66 & 0.54 & 0.46 & 0.73 \\
  $\ast$ FedCLIP$^\ddag$ \cite{lu2023fedclip} & 41.76 & 0.42 & 0.45 & 0.41 & 0.88 & 0.61 & 85.25 & 0.85 & 0.85 & 0.85 & 0.84 & 0.90 & 56.14 & 0.59 & 0.64 & 0.56 & 0.44 & 0.72 \\
    \rowcolor{gray!15}$\ast$ FedRep \cite{collins2021exploiting} & 80.61 & 0.75 & 0.81 & 0.71 & 0.90 & 0.91 & 94.13 & 0.94 & 0.94 & 0.94 & 0.94 & 0.97 & 83.27 & 0.74 & 0.79 & 0.71 & 0.96 & 0.93 \\
 $\ast$ FedDM \cite{xiong2023feddm} & 79.81 & 0.75 & 0.79 & 0.73 & 0.92 & 0.92 & 94.47 & 0.94 & 0.94 & 0.94 & 0.95 & 0.98 & 82.27 & 0.75 & 0.79 & 0.73 & 0.95 & 0.93 \\
  \rowcolor{gray!15}$\ast$ FedMCA \cite{zheng2025personalized} & 80.06 & 0.75 & 0.80 & 0.72 & 0.94 & 0.93 & 90.85 & 0.90 & 0.91 & 0.90 & 0.95 & 0.97 & 81.82 & 0.74 & 0.78 & 0.70 & 0.92 & 0.92 \\ 
    $\ast$ FtMoE \cite{liu2025ftmoe} & 61.75 & 0.53 & 0.56 & 0.52 & 0.90 & 0.83 & 90.14 & 0.90 & 0.90 & 0.89 & 0.95 & 0.98 & 78.06 & 0.65 & 0.68 & 0.63 & 0.93 & 0.93 \\ 
      \rowcolor{gray!15} Swin-B \cite{liu2021swin} & \textcolor{cyan}{84.94} & \textcolor{cyan}{0.84} & \textcolor{cyan}{0.84} & \textcolor{cyan}{0.84} & 0.95 & \textcolor{cyan}{0.96} & 93.95 & 0.93 & 0.93 & 0.93 & 0.97 & 0.98 & 85.46 & 0.85 & 0.85 & 0.85 & 0.96 & 0.97 \\
    CoAtNet \cite{dai2021coatnet} & 77.19 & 0.71 & 0.74 & 0.69 & 0.91 & 0.93 & 91.35 & 0.91 & 0.91 & 0.91 & 0.97 & 0.97 & 74.84 & 0.64 &0.65 & 0.63 & 0.90 & 0.94 \\
     \rowcolor{gray!15} MixNet \cite{tan2019mixconv} & 81.73 & 0.78 & 0.79 & 0.77 & \textcolor{cyan}{0.95} & 0.95 & 94.79 & 0.94 & 0.94 & 0.94 & 0.94 & 0.99 & 83.97 & 0.76 & 0.80 & 0.73 & 0.96& 0.96 \\
    $\ast$ DAFL & 83.45 & 0.80 & 0.83 & 0.79 & 0.94 & 0.95 & \textcolor{cyan}{96.63} & \textcolor{cyan}{0.96} & \textcolor{cyan}{0.96} & \textcolor{cyan}{0.96} & \textcolor{cyan}{0.98} & \textcolor{cyan}{0.99} & \textcolor{cyan}{87.19} & \textcolor{cyan}{0.87}  & \textcolor{cyan}{0.86} & 0.81 & \textcolor{cyan}{0.97} & \textcolor{cyan}{0.97} \\
  
    \bottomrule
  \end{tabular}
  \vspace{1mm}
  {$\dag$, and $\ddag$ represent ViT-B/16 and ViT-B/32 image encoder resepctively. $\ast$ represent the FL methods, without $\ast$ represents the centralized methods.} 
  \label{Tab:Medical}\vspace{-0.5cm}
\end{table*}

\subsection{Implementation details}
We considered ViT-S/16 as the backbone derived from \cite{rwightman2019timm}; its model size is suitable for the FL environment, which requires the computational resources of multiple local clients. We evaluated the most representative CNNs, including ResNet50 \cite{he2016deep} and DenseNet121 \cite{huang2017densely}. Additionally, we considered ViT-based models and their variants, such as ViT-S/16 \cite{rwightman2019timm} and ViT-L/32 \cite{rwightman2019timm}, along with the Swin Transformer-B (Swin-B) \cite{liu2021swin}. In addition, we studied models that integrate CNN and / or ViT characteristics, such as CoAtNet \cite{dai2021coatnet}, MixNet \cite{tan2019mixconv}. {We evaluate five state-of-the-art FL methods—FedCLIP \cite{lu2023fedclip}, FedRep \cite{collins2021exploiting}, FedDM \cite{xiong2023feddm}, FedMCA \cite{zheng2025personalized}, and FtMoE \cite{liu2025ftmoe}.Specifically, FedCLIP (Vision-language-based model), FedRep (representation-classifier decoupling), FedMCA (statistical heterogeneity calibration), FtMoE (architectural Mixture of expert innovation), and FedDM (data-level distribution matching). By spanning pre-trained models, representation learning, and data synthesis, this comprehensive benchmark allows for a multi-dimensional evaluation of the proposed method to address dual imbalance challenges.} We consistently used a learning rate of $1\times 10^{-4}$, a batch size of 16, and applied the Adam optimizer to all models \cite{kingma2014adam}. \textit{Our experiments were conducted on a Windows 11 system, equipped with a Nvidia RTX 4090 GPU, using PyTorch 1.8.1. The random seed is set to 0 for a fair comparison.}

For federated settings, each communication round involved training the local model for a single epoch, and the total number of communication rounds was set to 50. For FL methods, we split the training data from each data set into three separate clients (denoted as $C_{1},...,C_{3}$), while the test set was designated as the global test set (denoted as $C_{glo}$). In centralized deep learning techniques, the entire training dataset is used without dividing it among clients, and the test dataset is reserved exclusively for evaluation purposes. For all data sets in the centralized learning experiments, we split the training set randomly into training, validation using a 9:1 ratio.

Furthermore, data are often distributed unevenly among clients, known as non-IID data. To replicate this condition, we implemented a Dirichlet distribution strategy to divide datasets among clients. The Dirichlet distribution produces a series of random values that sum to 1, which we leverage as the basis for determining the data share for each client. By tuning the Dirichlet distribution concentration parameter, we can control the degree of data imbalance: Lower $\beta$ values result in highly biased data assignments (more imbalance), whereas higher values yield a more equal data distribution. This technique is applied to assign datasets to clients within the FL model.

The Dirichlet distribution can be mathematically represented as:
\begin{equation}
    \mathbf{q} \sim \text{Dirichlet}(\beta \times \mathbf{e}_M)
    \label{dirichlet}
\end{equation}
where \(\mathbf{q} = (q_1, q_2, \ldots, q_M)\) denotes the proportions of data assigned to each of the \(M\) clients. The parameter \(\beta\) determines the concentration of the distribution, and \(\mathbf{e}_M\) is a vector of lengths \(M\), representing the same initial weights for all clients. In practice, the ratios \(\mathbf{q}\) can be generated from the Dirichlet distribution or predefined according to specific rules. For example, a fixed ratio array like \([0.4, 0.3, 0.2, 0.1]\) would allocate 40\%, 30\%, 20\%, and 10\% of the data to the clients, respectively. For each class \(c\) in the dataset, the number of samples given to the client \(j\), represented as \(S_j\), is calculated by multiplying the ratio \(q_j\) by the total number of samples in class \(c\), represented as \(N_c\). The ratios we used were \([0.5, 0.3, 0.2]\). Table \ref{tab:Samples} reports the number of samples distributed to each client in different datasets. To simulate non-IID data distributions in the FL setting, we employ the Dirichlet distribution see (Eq. \ref{dirichlet}) to generate data proportions for each client. However, for the sake of experimental reproducibility and variable control, fixed ratio arrays (such as [0.5, 0.3, 0.2]) are used for actual data partitioning. This fixed ratio can be regarded as a deterministic special case of the Dirichlet distribution under specific concentration parameters.


We use Accuracy, Precision, Recall, F1 Score, Specificity, and AUC to assess the model performance. In addition, the clinical advantages of these methods are evaluated through Decision Curve Analysis (DCA) \cite{vickers2019simple}.

\begin{figure}[]
     \centering
\includegraphics[width=0.9\linewidth]{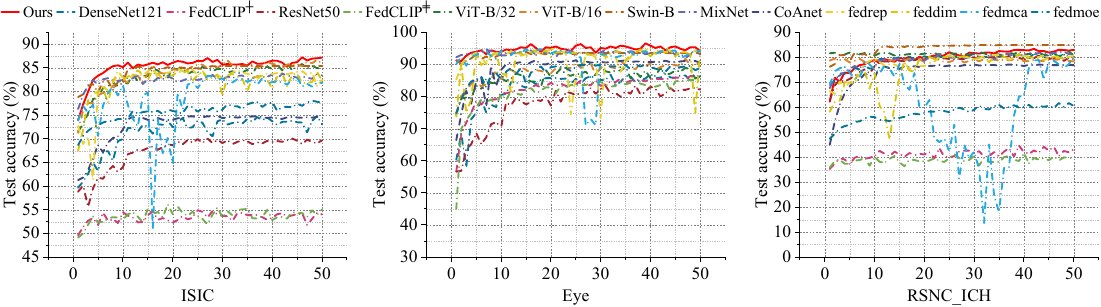}
    \caption{{Classification accuracy for each epoch using test samples of Ocular disease, ISIC, and RSNA-ICH.}}
    \label{x}
\end{figure}

\begin{figure}[]
     \centering
\includegraphics[width=0.9\linewidth]
{ 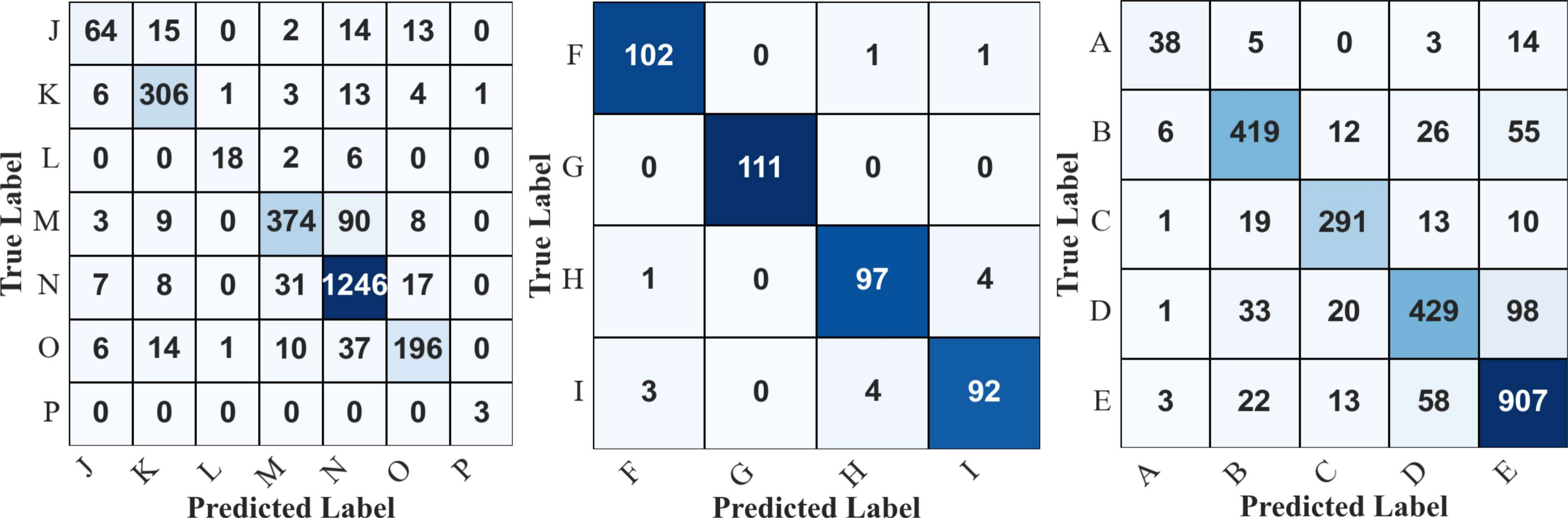}
    \caption{The confusion matrices obtained by DAFL on ISIC (\textbf{Left}),  Ocular Disease (\textbf{Middle}), and RSNA-ICH (\textbf{Right}).}
    \label{confusion_matrix}
\end{figure}

\subsection{Results} 
\noindent\mypar{RSNA-ICH} As reported in Table \ref{Tab:Medical}, DAFL provide an accuracy of 83.45\%, outperforming the accuracy of ViT-S/16 (79.87\%), ViT-L/32 (82.47\%), FedCLIP (incorporating ViT-B/16 and ViT-B/32 backbones), ResNet50 (79.53\%), and CoAtNet (77.19\%), while demonstrating comparable performance to DenseNet121 (81.26\%) and MixNet (81.73\%). {Furthermore, DAFL shows a better test accuracy compared to heterogeneity-aware FL methods such as FedRep. For example, it outperforms FedRep (80.61\%) and FedMCA (80.06\%) by approximately 2.8\% and 3.4\%, respectively. This suggests that the proposed dynamic loss adjustment is effective at handling class imbalance in CT scans than decoupling representations (FedRep) or knowledge distillation (FedMCA). However, Swin-B (84.94\%) provides a higher accuracy than DAFL. This performance gap might be due to the capability of Swin Transformer in capturing multi-scale features relative to the conventional ViT backbone.}

\noindent\mypar{Ocular Disease} Again, DAFL achieved an accuracy of 96.63\%, surpassing all benchmark models as reported in Table \ref{Tab:Medical}. The comparative performances are as follows: ViT-S/16 at 91.00\%, ViT-L/32 at 92.00\%, FedCLIP$^\dag$ at 86.42\%, FedCLIP$^\ddag$ at 85.25\%, DenseNet121 at 87.07\%, Swin-B at 93.95\%, CoAtNet at 91.35\%, MixNet at 94.79\%, and ResNet50 at 82.21\%. {Notably, DAFL outperforms the FedDM (94.47\%) and FedRep (94.13\%). These results suggest that DAFL preserves original feature information through adaptive weighting, while the synthetic data used in FedDM may lose fine-grained details of retinal lesions.}

\noindent\mypar{ISIC} As listed in Table \ref{Tab:Medical}, DAFL outperformed most other models, achieving an accuracy of 87.19\%, which is higher than ViT-S/16 (83.77\%), ViT-L/32 (85.84\%), DenseNet121 (74.31\%), ResNet50 (69.09\%), Swin-B (85.46\%), CoAtNet (74.84\%), and MixNet (83.97\%). {In addition, DAFL outperforms FtMoE (78.06\%) and FedCLIP variants with a large margin (e.g., +9.13\%). This suggests that MoE or adapter structures may struggle to converge with limited local data in highly imbalanced skin lesion datasets.}

Figure \ref{x} shows the global test accuracy for each epoch. Based on the experimental results presented in Table \ref{x}, DAFL demonstrates strong competitiveness in multiple data sets. While ViT-S/16 and ViT-L/32 demonstrated solid results on select datasets, DAFL showed quicker convergence, attaining high accuracy with a reduced number of communication rounds. For instance, in the RSNA-ICH datasets, DAFL achieved similar or superior accuracy with fewer communication epochs, underscoring the potential of DAFL to enhance training efficiency and lower communication costs. In summary, across a range of evaluation metrics, including F1 score, precision, recall, and AUC, DAFL demonstrated consistently high performance. Notably, in the Ocular Disease dataset, it attained an AUC of 0.99, which is markedly superior to the FedCLIP variants (0.84 and 0.85) and comparable to the results achieved by ViT-S/16 and ViT-L/32. Moreover, DAFL effectively balanced specificity with recall. 

\noindent\mypar{Confusion matrix.} In Figure \ref{confusion_matrix}, the confusion matrix for the three datasets is presented. Referring to Table \ref{table:class_name}, classes N, G, and E are the most accurately classified in the ISIC, Ocular Disease, and RSNA-ICH datasets, respectively. Moreover, for rare cancers such as J, L, and P in the ISIC dataset, the proposed model achieves a high classification rate (e.g., 64 out of 108 samples correctly classified for class J). These findings highlight the effectiveness of the adaptive focal loss employed in the proposed approach.

\begin{figure}[]
     \centering
\includegraphics[width=1\linewidth]{ 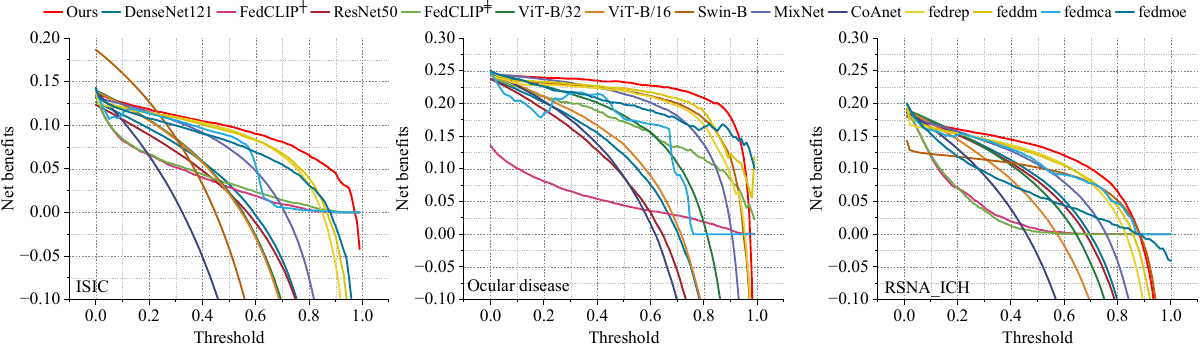}
    \caption{{Decision Curve Analysis (DCA) comparing the net benefit of various models across different thresholds for three datasets (ISIC, Ocular Disease, and RSNA-ICH). The x-axis represents the threshold probability, while the y-axis shows the net benefit.}}
    \label{DCA}
\end{figure} 

\begin{figure}[]
     \centering
\includegraphics[width=1\linewidth]{ 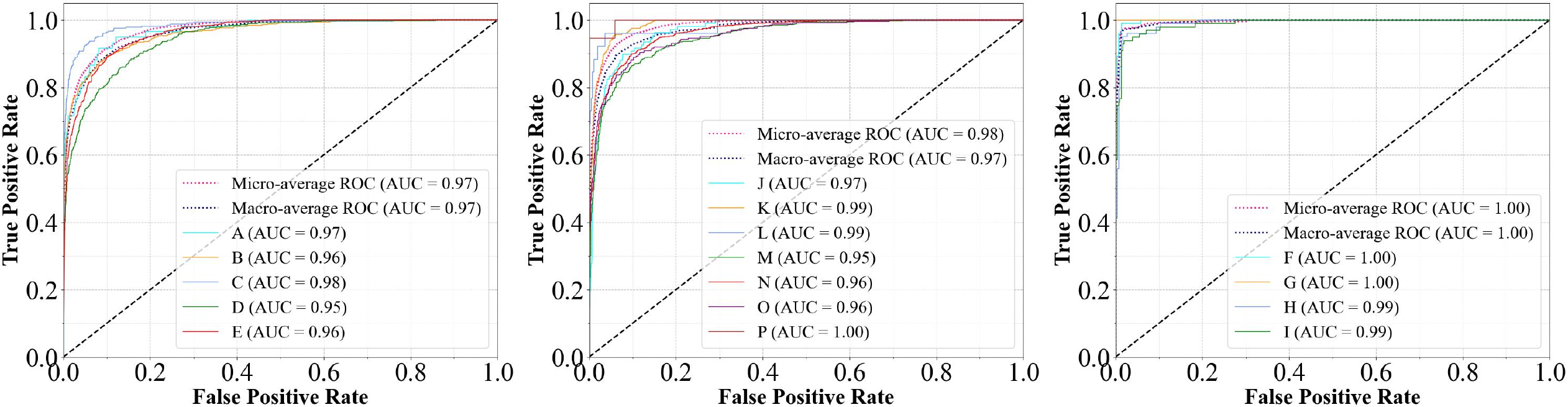}
    \caption{{ROC curves of ISIC, Ocular Disease, and RSNA-ICH. The performance of individual classes is evaluated using AUC (The value in bracket).}}
    \label{roc}
\end{figure}

\noindent\mypar{Decision curve analysis.} Figure \ref{DCA} shows the DCA curves for the 10 models. For ISIC, the proposed model outperforms others at low thresholds (e.g., 0.2-0.4), maintaining a net benefit of approximately 0.15–0.25, significantly higher than DenseNet121 (around 0.05–0.1). For ocular disease, the proposed model also shows strong performance in mid-thresholds (0.4-0.6), achieving a net benefit above 0.3, while Swin-B and MixNet remain around 0.2-0.25. For RSNA-ICH, the proposed model leads at high thresholds (0.6–0.8), with a net benefit of 0.1–0.2, compared to ResNet50 and ViT-B/16, which drop below 0.1. These findings highlight the capacity of the proposed model to provide important clinical value in various situations and thresholds. 

\noindent\mypar{ROC curves.} Figure \ref{roc} shows the ROC curves and AUC values for the three datasets. For example, The ISIC and RSNA-ICH datasets demonstrate consistently high AUC values ranging from 0.95 to 1.00. Similarly, the Ocular Disease dataset achieves an AUC value of 1.00 for multiple classes. These results suggest that DAFL provides remarkable performance for rare and common class samples.

\noindent\mypar{Computational overhead.}
Table \ref{tab:TotalTrainingCost} reports the computational time per second to transform the model in the classification of medical images using centralized and FL methods (i.e., FedCLIP$^\ddag$). For example, MixNet shows the minimum convergence time of 2678.88 s, compared to DAFL (our) with 3255.08 s using the RSNA-ICH dataset. However, for the Ocular Disease dataset, DAFL exhibits optimal training efficiency with a cost of 760s, substantially lower than DenseNet121 (926.9 s), ResNet50 (1277.64 s), and Swin-B (1105.16 s). For the ISIC dataset, while FedCLIP$^\ddag$ has the lowest cost of 7014.96 s, the training time of DAFL of 10563.9 s remains competitive given its superior predictive capabilities. We find that no single model uniformly excels in terms of training cost. However, DAFL demonstrates feasible efficiency striking a commendable balance between computational efficiency and predictive performance, as indicated in Table~\ref{Tab:Medical}. 

\begin{table}[ht!]\scriptsize
  \centering
  \caption{{Computational cost for training centralized and federated learning models (Epochs $\times$ Time per Epoch (s)) for classification tasks. * indicates the federated methods.}}
  \renewcommand{\arraystretch}{0.8}
  \setlength{\tabcolsep}{13pt}
  \begin{tabular}{lccc}
    \toprule
    \textbf{Model} & \textbf{RSNA-ICH} & \textbf{Ocular Disease} & \textbf{ISIC} \\
    \midrule
    \rowcolor{gray!15}DenseNet121 \cite{huang2017densely}  &4030.00  &926.90   &15549.12  \\
    ResNet50 \cite{he2016deep}    &4468.56  &1277.64  &9966.30  \\
    \rowcolor{gray!15}ViT-S/16 \cite{rwightman2019timm}      &2867.04  &1145.15  &8283.90  \\
    ViT-L/32 \cite{rwightman2019timm}      &7457.70  &1350.44  &17619.42  \\
    \rowcolor{gray!15}$\ast$ FedCLIP$^\dag$ \cite{lu2023fedclip}   &4031.20  &1618.20  &9827.21  \\
    $\ast$ FedCLIP$^\ddag$ \cite{lu2023fedclip}  
    &7544.52  &1239.68  &\textcolor{cyan}{7014.96}  \\
    \rowcolor{gray!15}$\ast$ FedMCA \cite{zheng2025personalized}  
    &7304.00  &2802.10  &17060.76  \\
    $\ast$ FedRep \cite{collins2021exploiting}  
    &5070.00  &2699.72  & 12828.32\\
    \rowcolor{gray!15}$\ast$ FedDM \cite{xiong2023feddm}  
    &3066.59  &805.48  &10381.67  \\
    $\ast$ FtMoE \cite{liu2025ftmoe}  
    &2809.85  &826.00  &8993.87  \\
    \rowcolor{gray!15}Swin-B \cite{liu2021swin}           &3536.91  &1105.16  &8560.03  \\
    CoAtNet \cite{dai2021coatnet}         &3104.08  &905.75  &9708.90  \\
    \rowcolor{gray!15}MixNet \cite{tan2019mixconv}          &\textcolor{cyan}{2678.88}  &\textcolor{cyan}{680.80} &14220.78  \\
    $\ast$  DAFL       &3255.08  &{760.00}  &10563.90  \\
    \bottomrule
  \end{tabular}
  \label{tab:TotalTrainingCost}
\end{table}

\noindent\mypar{Interpretability analysis}
To explore the interpretability of deep models, we use gradient-based attention rollout \cite{vit_rollout} to explain model prediction.  In gradient-based attention rollout, each attention layer produces a map \(A^{(l)}\) together with its gradient \(G^{(l)}\). We fuse these by computing the following equation: 
\begin{equation}
    F^{(l)} = \operatorname{ReLU}\bigl(A^{(l)} \odot G^{(l)}\bigr)
\end{equation}

\begin{figure}[]
     \centering
\includegraphics[width=0.98\textwidth]{ 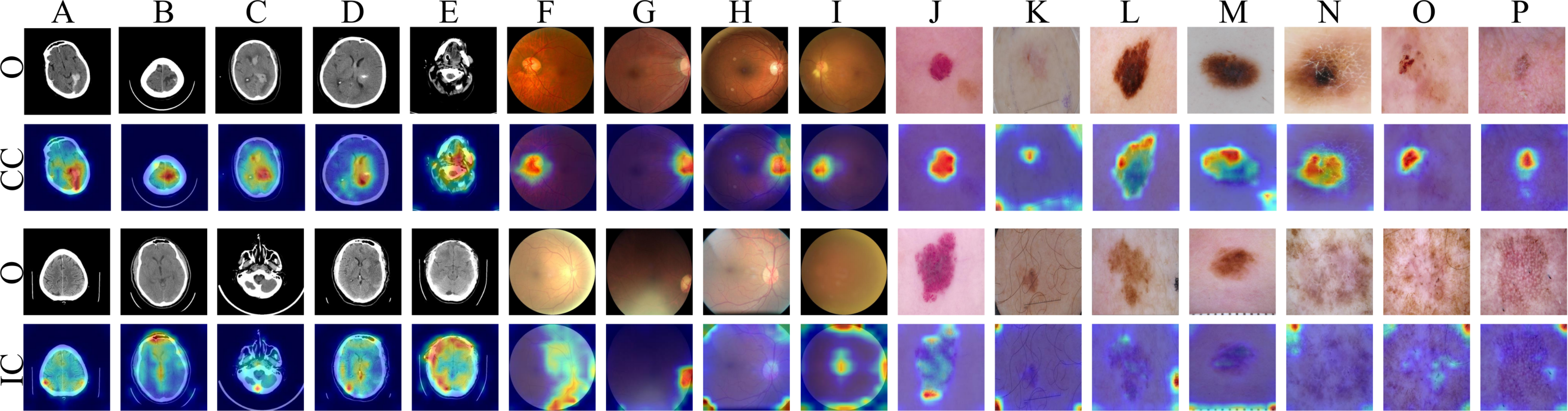}
    \caption{Visualization results using the XAI (Gradient-based rollout) method on three medical datasets. 'O', 'CC', and 'IC' indicate the original image, correctly classified samples, and incorrectly classified samples, respectively. 'A' to 'P' represent the class names.}
    \label{xai111}
\end{figure}

Then, we propagate the normalized maps through all layers, i.e., \(R = \prod_{l=1}^{L} \tilde{A}^{(l)}\), where the final mask \(M = R_{0,\,1:N}\) highlights the influential regions. This gradient-based rollout explains the predictions by revealing the input areas with the highest sensitivity. Figure \ref{xai111} illustrates the visualization heatmap obtained using gradient attention rollout. For correctly classified samples, for the RSNA-ICH dataset, DAFL can highlight the correct abnormal regions. Similar to RSNA-ICH, DAFL provides reasonable interpretability for skin cancer diagnosis (e.g., it focuses on the cancer region of class ``J''). However, for misclassified samples, for RSNA-ICH dataset, DAFL demonstrates dispersed attention across the image, as illustrated by intracranial hemorrhage in class `E'. We argue that this is because the abnormal region is not clearly visible in the image (e.g., class ``A''), preventing the model from correctly shifting its attention to the abnormal area. Similar conclusion can be drawn from ISIC dataset (e.g., class ``N'').

\subsection{Model analysis}

\noindent\mypar{FL-based techniques.} We compared DAFL with a centralized approach and other FL methods, including MOON \cite{li2021model} and FedProx \cite{li2020federated} while maintain the same hyperparameter setting. Table \ref{ab:FL} summarizes the test metrics in three datasets using centralized and federated frameworks. The results indicate that while centralized training, FedProx and MOON have lower accuracy compared to DAFL, their recall rates on the ISIC dataset are higher (e.g. 0.85 for MOON vs 0.81 for DAFL). This suggests that employing more advanced FL techniques could further enhance the overall performance of DAFL.

\begin{table*}[!ht]\scriptsize
  \centering
  \caption{Results of the ablation study on FL models. * indicates the federated methods}
  \renewcommand{\arraystretch}{0.7}  
  \setlength{\tabcolsep}{4.0pt}  
  \begin{tabular}{ccccccc|cccccc|cccccccccc}
    \toprule
    & \multicolumn{3}{c} \textbf{Accuracy} &\multicolumn{3}{c} \textbf{F1} & \multicolumn{3}{c}\textbf{Precision} & \multicolumn{3}{c}\textbf{Recall} &\multicolumn{3}{c} \textbf{Specificity} & \multicolumn{2}{c}\textbf{AUC}\\
    \cmidrule(lr){2-19}
    \multirow{1}{*}{Model} & \multicolumn{6}{c|}{RSNA-ICH} & \multicolumn{6}{c|}{Ocular Disease} & \multicolumn{6}{c}{ISIC} \\
    \midrule
      \rowcolor{gray!15}$\ast$ MOON\cite{li2021model} & 80.65 & 0.80 & 0.80 & 0.80 & 0.93 & 0.93 & 95.67 & 0.95 & 0.95 & 0.95 & 0.94 & 0.99 & 84.93 & 0.84 & 0.84 & 0.85 & 0.96 & 0.96 \\$\ast$ FedProx\cite{li2020federated} & 81.85 & \textcolor{cyan}{0.81} & 0.81 & \textcolor{cyan}{0.81} & 0.94 & 0.93 & 96.63 & 0.96 & 0.96 & 0.96 & 0.98 & 0.99 & 85.64 & 0.85 & 0.85 & \textcolor{cyan}{0.85} & 0.96 & 0.96 \\
    \rowcolor{gray!15} Centralized & 80.29 & 0.80 & 0.80 & 0.80 & 0.93 & 0.86 & 94.71 & 0.94 & 0.94 & 0.94 & 0.95 & 0.96 & 83.47 & 0.83 & 0.83 & 0.83 & 0.96 & 0.97 \\
     $\ast$ DAFL & \textcolor{cyan}{83.45} & 0.80 & \textcolor{cyan}{0.83} & 0.79 & \textcolor{cyan}{0.94} & \textcolor{cyan}{0.95} & \textcolor{cyan}{96.63} & \textcolor{cyan}{0.96} & \textcolor{cyan}{0.96} & \textcolor{cyan}{0.96} & \textcolor{cyan}{0.98} & \textcolor{cyan}{0.99} & \textcolor{cyan}{87.19} & \textcolor{cyan}{0.86} & \textcolor{cyan}{0.86} & 0.81 & \textcolor{cyan}{0.97} & \textcolor{cyan}{0.97} \\
    \bottomrule
  \end{tabular}
  \vspace{1mm}

{$\ast$ represent the FL methods, without $\ast$ represents the centralized methods.}
  \label{ab:FL}
\end{table*}

\noindent\mypar{Loss functions.} We performed controlled ablations comparing standard CE, conventional focal loss, and the proposed DAFL in RSNAICH, Ocular Disease, and ISIC while maintaining the same hyperparameter setting. As reported in Table .\ref{Tab:Medical_Extension} the experiments show consistent, data‑driven advantages of DAFL, DAFL achieves the highest accuracy on RSNA‑ICH (83.45\%) versus focal (83.17\%) and CE (80.85\%), and it attains the best F1 on ISIC (0.83) compared to focal loss (0.82) and CE (0.79). These improvements are consistent across datasets and most pronounced on imbalanced sets (ISIC), indicating better minority‑class handling. In addition, DAFL converges as faster than alternatives in our federated setup. For example, DAFL achieves the highest training efficiency on datasets RSNA-ICH and ISIC.

\begin{table*}[!ht]\scriptsize
  \centering
  \caption{Ablation study comparing DAFL with cross-entropy (CE) and standard focal loss. Computational cost for Focal loss and DAFL(Epochs $\times$ Time per Epoch (s)) for classification tasks. }
  \renewcommand{\arraystretch}{0.7}  
  \setlength{\tabcolsep}{2.0pt}  
  \begin{tabular}{c|ccccccc|ccccccc|ccccccc}
    \toprule
    & \multicolumn{3}{c}{\textbf{Accuracy}} & \multicolumn{3}{c}{\textbf{F1}} & 
      \multicolumn{3}{c}{\textbf{Precision}} & \multicolumn{3}{c}{\textbf{Recall}} &
      \multicolumn{3}{c}{\textbf{Specificity}} & \multicolumn{2}{c}{\textbf{AUC}} & \multicolumn{4}{c}{\textbf{Computational cost}} \\
    \cmidrule(lr){2-22}
    \multirow{1}{*}{Model} & \multicolumn{7}{c|}{RSNA-ICH} & \multicolumn{7}{c|}{Ocular Disease} & \multicolumn{7}{c}{ISIC} \\
    \midrule
    
    \rowcolor{gray!15}CE & 80.85 & 0.75 & 0.80 & 0.73 & 0.93 & 0.93 & 4363.20 & 
        95.43 & 0.95 & 0.95 & 0.95 & 0.95 & 0.98 & 753.7 & 
        85.42 & 0.80 & 0.83 & 0.79 & 0.96 & 0.96 & 12389.6 \\
    Focal loss & 83.17 & 0.79 & 0.83 & \textcolor{cyan}{0.80} & 0.94 & 0.95 & 3536.4 & 
        96.04 & 0.95 & 0.95 & 0.95 & 0.96 & 0.98 & \textcolor{cyan}{594.56} & 
        86.79 & 0.82 & 0.86 & 0.81 & 0.97 & 0.96 & 11589.26 \\
    \rowcolor{gray!15}DAFL & \textcolor{cyan}{83.45} & \textcolor{cyan}{0.80} & \textcolor{cyan}{0.83} & 0.79 & \textcolor{cyan}{0.94} & \textcolor{cyan}{0.95} & \textcolor{cyan}{3255.08} & 
        \textcolor{cyan}{96.63} & \textcolor{cyan}{0.96} & \textcolor{cyan}{0.96} & \textcolor{cyan}{0.96} & \textcolor{cyan}{0.96} & \textcolor{cyan}{0.98} & 760 & 
        \textcolor{cyan}{87.19} & \textcolor{cyan}{0.83} & \textcolor{cyan}{0.86} & \textcolor{cyan}{0.81} & \textcolor{cyan}{0.97} & \textcolor{cyan}{0.97} & \textcolor{cyan}{10563.9} \\
    \bottomrule
  \end{tabular}
  \label{Tab:Medical_Extension}
\end{table*}

\begin{table*}[h]\scriptsize
  \centering
  \caption{Performance metrics (\%) for three data distribution approaches. C1-C3 indicates three different data distribution. }
  \renewcommand{\arraystretch}{0.8}  
  \setlength{\tabcolsep}{4.5pt}  
  \begin{tabular}{ccccccc|cccccc|cccccccccc}
    \toprule
    & \multicolumn{3}{c} \textbf{Accuracy} &\multicolumn{3}{c} \textbf{F1} & \multicolumn{3}{c}\textbf{Precision} & \multicolumn{3}{c}\textbf{Recall} &\multicolumn{3}{c} \textbf{Specificity} & \multicolumn{2}{c}\textbf{AUC}\\
   \cmidrule(lr){2-19}
   \multirow{1}{*}{Model} & \multicolumn{6}{c|}{RSNA-ICH} & \multicolumn{6}{c|}{Ocular Disease} & \multicolumn{6}{c}{ISIC} \\
    \midrule
     \rowcolor{gray!15}C1 & \textcolor{cyan}{83.45} & \textcolor{cyan}{0.80} & 0.83 & \textcolor{cyan}{0.79} & \textcolor{cyan}{0.94} & \textcolor{cyan}{0.95} & \textcolor{cyan}{96.63} & \textcolor{cyan}{0.96} & \textcolor{cyan}{0.96} & \textcolor{cyan}{0.96} & \textcolor{cyan}{0.98} & \textcolor{cyan}{0.99} & \textcolor{cyan}{87.19} & \textcolor{cyan}{0.83} & \textcolor{cyan}{0.86} & \textcolor{cyan}{0.81} & \textcolor{cyan}{0.97} & \textcolor{cyan}{0.97} \\
    C2 & 82.73 & 0.79 & \textcolor{cyan}{0.84} & 0.76 & 0.94 & 0.94 & 96.15 & 0.96 & 0.96 & 0.96 & 0.97 & 0.99 & 85.72 & 0.77 & 0.78 & 0.76 & 0.97 & 0.96 \\
     \rowcolor{gray!15}C3 & 81.05 & 0.76 & 0.82 & 0.72 & 0.94 & 0.93 & 95.91 & 0.95 & 0.95 & 0.95 & 0.97 & 0.99 & 79.83 & 0.76 & 0.74 & 0.77 & 0.96 & 0.93 \\
    \bottomrule
  \end{tabular}
  \label{ab:data_distribution}
\end{table*}

\begin{table*}[h]\scriptsize
  \centering
  \caption{Performance metrics (\%) using different learning rate.}
  \renewcommand{\arraystretch}{0.7}  
  \setlength{\tabcolsep}{4.5pt}  
  \begin{tabular}{ccccccc|cccccc|cccccccccc}
    \toprule
    & \multicolumn{3}{c} \textbf{Accuracy} &\multicolumn{3}{c} \textbf{F1} & \multicolumn{3}{c}\textbf{Precision} & \multicolumn{3}{c}\textbf{Recall} &\multicolumn{3}{c} \textbf{Specificity} & \multicolumn{2}{c}\textbf{AUC}\\
    \cmidrule(lr){2-19}
    \multirow{1}{*}{Model} & \multicolumn{6}{c|}{RSNA-ICH} & \multicolumn{6}{c|}{Ocular Disease} & \multicolumn{6}{c}{ISIC} \\
    \midrule
      \rowcolor{gray!15}\textbf{$1\times10^{-2}$} & 40.50 & 0.13 & 0.15 & 0.20 & 0.80 & 0.56 & 59.38 & 0.56 & 0.61 & 0.58 & 0.86 & 0.82 & 52.30 & 0.20 & 0.20 & 0.22 & 0.75 & 0.70 \\
    \textbf{$1\times10^{-3}$} & 37.46 & 0.17 & 0.13 & 0.22 & 0.81 & 0.58 & 83.41 & 0.82 & 0.83 & 0.83 & 0.94 & 0.94 & 71.41 & 0.50 & 0.63 & 0.52 & 0.80 & 0.93 \\
     \rowcolor{gray!15}\textbf{$1\times10^{-4}$} & \textcolor{cyan}{83.45} & \textcolor{cyan}{0.80} & \textcolor{cyan}{0.83} & \textcolor{cyan}{0.79} & \textcolor{cyan}{0.94} & \textcolor{cyan}{0.95} & \textcolor{cyan}{96.63} & \textcolor{cyan}{0.96} & \textcolor{cyan}{0.96} & \textcolor{cyan}{0.96} & 0.96 & 0.98 & \textcolor{cyan}{87.19} & \textcolor{cyan}{0.83} & \textcolor{cyan}{0.86} & \textcolor{cyan}{0.81} & \textcolor{cyan}{0.97} & \textcolor{cyan}{0.97}
     \\
    \textbf{$1\times10^{-5}$} & 79.69 & 0.74 & 0.79 & 0.71 & 0.94 & 0.93 & 94.95 & 0.94 & 0.94 & 0.94 & \textcolor{cyan}{0.98} & \textcolor{cyan}{0.99} & 85.23 & 0.75 & 0.80 & 0.72 & 0.97 & 0.97 \\
     \rowcolor{gray!15}\textbf{$1\times10^{-6}$} & 69.03 & 0.65 & 0.67 & 0.64 & 0.91 & 0.87 & 93.03 & 0.92 & 0.92 & 0.92 & 0.97 & 0.98 & 78.87 & 0.66 & 0.71 & 0.65 & 0.97 & 0.97 \\
    \textbf{$1\times10^{-7}$} & 48.28 & 0.41 & 0.40 & 0.43 & 0.86 & 0.72 & 81.49 & 0.81 & 0.81 & 0.81 & 0.92 & 0.92 & 66.55 & 0.30 & 0.46 & 0.30 & 0.71 & 0.71 \\
    \bottomrule
  \end{tabular}
  
  \label{ab:lr}
\end{table*}

\begin{table*}[!h]\scriptsize
  \centering
  \caption{Performance metrics (\%) using different batch size. }
  \renewcommand{\arraystretch}{0.8}  
  \setlength{\tabcolsep}{4.5pt}  
  \begin{tabular}{ccccccc|cccccc|cccccccccc}
    \toprule
    & \multicolumn{3}{c} \textbf{Accuracy} &\multicolumn{3}{c} \textbf{F1} & \multicolumn{3}{c}\textbf{Precision} & \multicolumn{3}{c}\textbf{Recall} &\multicolumn{3}{c} \textbf{Specificity} & \multicolumn{2}{c}\textbf{AUC}\\
    \cmidrule(lr){2-19}
    \multirow{1}{*}{Batch size} & \multicolumn{6}{c|}{RSNA-ICH} & \multicolumn{6}{c|}{Ocular Disease} & \multicolumn{6}{c}{ISIC} \\
    \midrule
      \rowcolor{gray!15}4 &70.65 & 0.63 & 0.64 & 0.63 & 0.91 & 0.84 & 94.10 & 0.94 & 0.94 & 0.94 & 0.98 & 0.98 & 76.45 & 0.65 & 0.66 & 0.66 & 0.94 & 0.88 \\
    8 &75.08 & 0.67 & 0.68 & 0.67 & 0.92 & 0.87 & 93.87 & 0.93 & 0.93 & 0.93 & 0.97 & 0.98& 80.64 & 0.73 & 0.75 & 0.72 & 0.96 & 0.94 \\
     \rowcolor{gray!15}16 &\textcolor{cyan}{83.45} & \textcolor{cyan}{0.80} & \textcolor{cyan}{0.83} & \textcolor{cyan}{0.79} & \textcolor{cyan}{0.94} & \textcolor{cyan}{0.95} & \textcolor{cyan}{96.63} & \textcolor{cyan}{0.96} & \textcolor{cyan}{0.96} & \textcolor{cyan}{0.96} & \textcolor{cyan}{0.98} & \textcolor{cyan}{0.99} & \textcolor{cyan}{87.19} & \textcolor{cyan}{0.83} & \textcolor{cyan}{0.86} & \textcolor{cyan}{0.81} & \textcolor{cyan}{0.97} & \textcolor{cyan}{0.97} \\
    32 &77.00 & 0.69 & 0.71 & 0.68 & 0.93 & 0.91 & 95.19 & 0.95 & 0.95 & 0.95 & 0.98 & 0.99 & 85.64 & 0.79 & 0.81 & 0.78 & 0.96 & 0.96 \\
     \rowcolor{gray!15}64 &79.13 & 0.73 & 0.77 & 0.71 & 0.94 & 0.93 & 96.09 & 0.95 & 0.95 & 0.95 & 0.98 & 0.99 & 86.35 & 0.79 & 0.80 & 0.81 & 0.97 & 0.97 \\
    \bottomrule
  \end{tabular}
  \label{ab:batch_size}
\end{table*}

\noindent\mypar{Data distribution}
We used the Dirichlet distribution with a concentration vector of [0.5,0.3,0.2], denoted as C1, in our primary experimental setup to simulate data imbalance among clients. We also considered two alternative concentration vectors: [0.333, 0.333, 0.333], represented as C2, and [0.556, 0.278, 0.166], represented as C3. The results are reported in Table \ref{ab:data_distribution}. For the RSNA-ICH dataset, distribution C1 achieves the highest accuracy of 83.45\% and an F1 of 0.80, indicating a balanced model with strong precision and recall. In the Ocular Disease dataset, C1 also performs the best, with an accuracy of 96.63\% and an F1 of 0.96, effectively capturing data complexity while maintaining high precision and recall. For the ISIC dataset, C1 also provides better results, with an accuracy of 87.19\% and an F1 of 0.83. However, other distributions are not performing well, particularly in precision and recall. These results suggest stable performance of DAFL with increasing data heterogeneity for the RNSA-ICH and Ocular Disease datasets but indicate decreased performance for the ISIC dataset. To improve the performance of the ISIC dataset, further improvements will be considered as suggested in \cite{mavaddati2025skin}. We attribute the superior performance of C1 to clients with larger data allocations providing more stable and representative gradient updates. During global aggregation, $c_{f, \text{client}}$ assigns higher weights to such clients, ensuring that their feedback dominates the global update.

\noindent\mypar{Learning rate.}
Table \ref{ab:lr} reports the impact of different learning rates (LR) on the RSNA-ICH, Ocular Disease, and ISIC datasets. For all datasets, the LR of $1 \times 10^{-4}$ consistently provides the best performance. The three datasets achieve accuracy and AUC values of 83.45\%-0.94 for RSNA-ICH, 96.63\%-0.98 for ocular diseases, and 87.19\%-0.97 for ISIC, respectively. Higher LR, such as $1 \times 10^{-2}$, leads to decrease in performance. These findings suggest that a proper LR is important, with $1 \times 10^{-4}$ proving to be optimal across all datasets.

\begin{table}[!ht]\scriptsize
\centering
\caption{{Summary of imbalance classes in datasets. The classes are sorted by sample size within each dataset. The ``Imbalance score'' indicates the rarity of the class.}}
\label{tab:dataset_stats}

\begin{tabular}{llccc}
\toprule
\textbf{Dataset} & \textbf{Class Name} & \textbf{Total samples} & \textbf{Imbalance score} & \textbf{Type} \\
\midrule
\multirow{5}{*}{\textbf{RSNA-ICH}} 
 & A & 9,126 & 1.4658 & Head \\
 & B  & 5,215 & 3.3151 & Head \\
 & C  & 4,644 & 3.8456 & Head \\
 & D  & 2,990 & 6.5261 & Tail \\
 & E  & 528 & 41.6193 & Tail \\
\midrule
\multirow{4}{*}{\textbf{Ocular Disease}} 
 & H  & 987 & 2.8399 & Head \\
 & F  & 965 & 2.9275 & Head \\
 & I  & 933 & 3.0622 & Head \\
 & G & 905 & 3.1878 & Tail \\
\midrule
\multirow{7}{*}{\textbf{ISIC}} 
 & L & 11,768 & 0.9432 & Head \\
 & M & 4,338 & 4.2713 & Head \\
 & P & 2,992 & 6.6427 & Head \\
 & K & 2,363 & 8.6771 & Head \\
 & O & 959 & 22.8446 & Head \\
 & J & 226 & 100.1814 & Tail \\
 & N & 221 & 102.4706 & Tail \\
\bottomrule
\label{tail/head}
\end{tabular}
\end{table}

\noindent\mypar{Batch size.} Table \ref{ab:batch_size} reports the test metrics for DAFL using different batch sizes. We observe that the model can provide higher test metrics (e.g., Accuracy of 83.45\% on RSNA-ICH, 96.63\% on Ocular Disease, and 0.83 F1 score on ISIC) when the batch size is set to 16. In addition, a small batch size (e.g., 4) can lead to a considerable performance decrease.

\begin{figure*}[]
    \centering
    \includegraphics[width=1.0\linewidth]{ 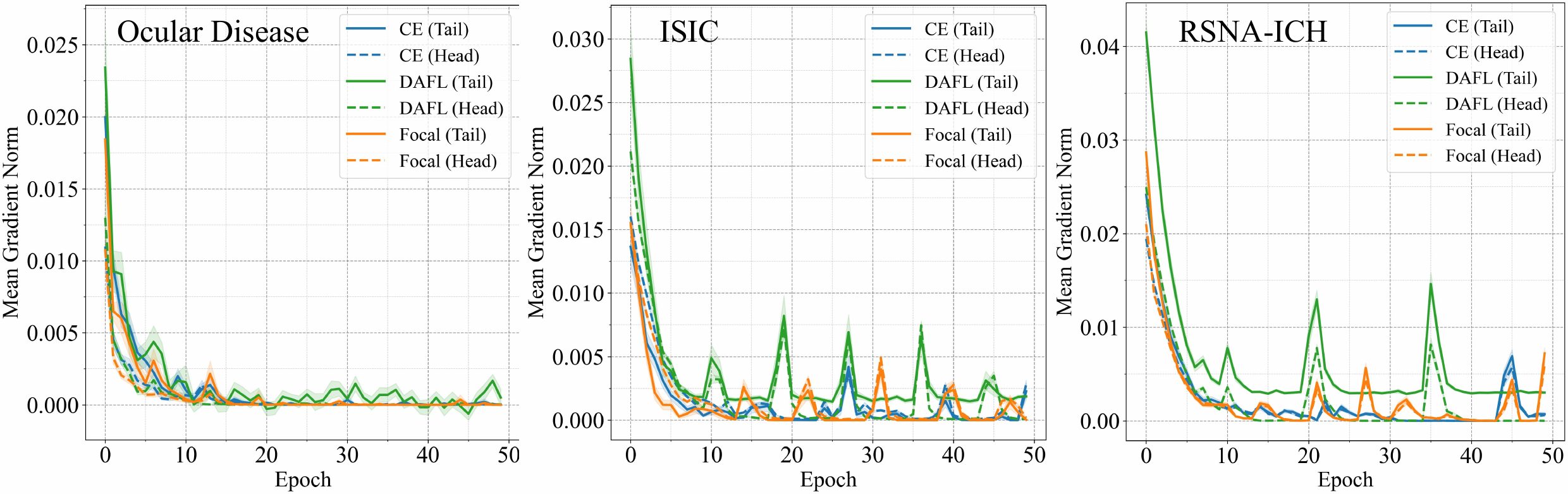} 
    \caption{{Analysis of gradient norms of classification across Ocular Disease, RSNA-ICH, and ISIC datasets: A comparison of mean gradients for CE, Focal Loss, and DAFL across Tail and Head classes (see Table \ref{tail/head} for class definitions).}}
    \label{fig:grad_norm}
\end{figure*}

\mypar{Gradient norm analysis} {Figure \ref{fig:grad_norm} shows the evolution of the mean gradient norms during training across the three datasets. We categorized the classes into "Tail" (the minority 30\% of classes) and "Head" (the majority 70\% of classes) as reported in Table \ref{tail/head}. To quantify the degree of class imbalance within the federated training data, we compute an imbalance score for each class. Let $N$ denote the total number of training samples across all participating clients, and let $n_k$ be the number of samples belonging to class $k$. The imbalance score $s_k$ for class $k$ is defined as $s_k = \frac{N - n_k}{n_k}$. The results demonstrate a distinct behavior in DAFL compared to CE and Focal Loss. Specifically, in the highly imbalanced ISIC and RSNA-ICH datasets, the gradient norms for DAFL (green lines) are consistently maintained at a higher magnitude than those of CE and Focal Loss (e.g., the dafl-tail curve (solid green) does not vanish rapidly). This result suggests that the dynamic coefficient $(1 + c_{f,t})$ effectively maintains a balance between majority and minority classes, ensuring that minority classes continue to contribute during optimization, preventing the model from overfitting to the head classes.}

\section{Discussion}
\label{discussion}

The experimental suite across RSNA-ICH, Ocular Disease and ISIC demonstrates three consistent outcomes, DAFL improves minority class sensitivity compared to standard cross entropy and focal loss, and the client aware aggregation reduces the dominance of large but internally skewed clients, yielding more stable federated convergence, and ViT backbones benefit from the dynamic reweighting, producing attention maps that better localize clinically relevant regions. It outperforms both centralized approaches and FL methods such as FedProx and MOON in various metrics, including accuracy, F1 score, specificity, and AUC. DAFL demonstrates robustness under heterogeneous data distributions, varying learning rates, and batch sizes, achieving over 80\% accuracy for the RSNA-ICH dataset in all federated data partitioning situations. For the ISIC dataset, despite a $\sim$7\% performance drop due to class imbalance, DAFL still outperforms the cross-entropy loss and baseline FL models. Its ability to handle unbalanced datasets highlights its potential to improve the scalability and robustness of FL systems in clinical applications. The effectiveness of DAFL in capturing representative features related to abnormal regions is consistent with previous findings that employ prototype learning to refine feature qualities \cite{chen2025prototype}. 

{Furthermore, the results on medical datasets (Table \ref{Tab:Medical}) demonstrate that simply using architectural decoupling, KD, image generation or expert aggregation showed limited performance improvements. These results can not capture dominant features in medical image (i.e., abnormal region represents a small part of the image). Unlike these methods, DAFL introduces re-weighting strategy and adaptive loss function to learn robust feature representations across clients, offering a more stable and universally applicable approach to handling heterogeneity.} Furthermore, DAFL results demonstrated feasible computational resources (Table \ref{tab:TotalTrainingCost} and Table \ref{Tab:Medical_Extension}). Notably, on the RSNA-ICH and ISIC, the DAFL and achieve the higher performance and time efficiency in RSNA-ICH and ISIC compared to the focal loss and standard CE.

The interpretability analyses using gradient attention rollout provide qualitative evidence that DAFL not only improves numeric metrics but also focuses model attention on lesion regions for correctly classified examples. This behavior supports clinical trustworthiness: better localization can aid human-in-the-loop verification and reduce the risk of spurious correlations. Nevertheless, attention maps for some misclassified or subtle cases remain diffuse, suggesting limits when the signal-to-noise ratio in an image is low.

\noindent\textit{\textbf{Limitations and future work.}} The experiments were limited to three datasets and a small number of simulated clients; and the aggregation rule relies on client-provided summary statistics, which may be noisy or manipulated in adversarial settings; (iii) we did not evaluate privacy-preserving mechanisms (e.g., DP or secure aggregation) that would interact with the statistics used by DAFL. Future work will integrate domain adaptation techniques as suggested in \cite{wu2024facmic} to align feature spaces and refine the aggregation algorithm with dynamic weighting strategies \cite{qi2024model} to improve fairness and performance in heterogeneous settings.

\section{Conclusion}\label{conclusion}
This paper presented dynamic adaptive focal loss to address data imbalance in the FL framework for medical image classification. DAFL method achieves superior classification accuracy, F1 score, and specificity compared to existing baseline and FL methods. Through ablation studies, we identified the optimal learning rate and batch size for DAFL  proposed method and validated the effectiveness of the proposed loss function compared to the standard CE loss. Our approach balances high performance with moderate computational overhead, a balance validated by cross-dataset analyses. Interpretability analysis using the Gradient Attention Rollout provides information on the model decision-making process, enhancing its clinical utility. However, feature biases caused by diversity in clients remain a limitation. Future work may focus on integrating domain adaptation techniques to reduce inter-client discrepancies and optimizing the federated aggregation algorithm to improve performance under diverse client conditions.

\section*{Acknowledgments}
This research was funded by the Guangxi Science and Technology Base and Talent Project (2022AC18004, 2022AC21040) and the National Natural Science Foundation of China grant number 82260360. 

{
\bibliographystyle{unsrt}
\bibliography{cas-refs}
}

\end{document}